\documentclass[journal]{IEEEtran}
\usepackage{bm}
\usepackage{amsmath}
\usepackage{amssymb}

\ifCLASSINFOpdf
  \usepackage[pdftex]{graphicx}
\else
  \usepackage[dvips]{graphicx}
\fi
\usepackage{algorithmic}
\usepackage{algorithm}
\begin{document}
%
\title{On the Matrix-Free Generation of Adversarial Perturbations for Black-Box Attacks}
%
%
%

\author{Hisaichi~Shibata,
        Shouhei~Hanaoka, 
        Yukihiro~Nomura, 
        Naoto~Hayashi,        
        and~Osamu~Abe
\thanks{H. Shibata, Y. Nomura, and N. Hayashi are with Department of Computational Diagnostic Radiology and Preventive Medicine, The University of Tokyo Hospital, 7-3-1 Hongo, Bunkyo-ku, Tokyo, 113-8655 Japan e-mail: sh-tky@umin.ac.jp.}
\thanks{S. Hanaoka and O. Abe are with Department of Radiology, The University of Tokyo Hospital, 7-3-1 Hongo, Bunkyo-ku, Tokyo, 113-8655 Japan.}
\thanks{O. Abe is with Division of Radiology and Biomedical Engineering, Graduate School of Medicine, The University of Tokyo, 7-3-1 Hongo, Bunkyo-ku, Tokyo, 113-8656 Japan}
}

\maketitle

\begin{abstract}
In general, adversarial perturbations superimposed on inputs are realistic threats for a deep neural network (DNN). 
In this paper, we propose a practical generation method of such adversarial perturbation to be applied to black-box attacks that demand access to an input-output relationship only.
Thus, the attackers generate such perturbation without invoking inner functions and/or accessing the inner states of a DNN.
Unlike the earlier studies, the algorithm to generate the perturbation presented in this study requires much fewer query trials.
Moreover, to show the effectiveness of the adversarial perturbation extracted, we experiment with a DNN for semantic segmentation.
The result shows that the network is easily deceived with the perturbation generated than using uniformly distributed random noise with the same magnitude.
\end{abstract}


%
\IEEEpeerreviewmaketitle

\section{Introduction}
\IEEEPARstart{T}{he} vulnerability of deep neural networks (DNNs) against a carefully generated small noise disturbance, i.e., \textit{adversarial perturbation}, superimposed on input data, is a common threat to be prevented \cite{papernot2016limitations,papernot2017practical,akhtar2018threat}.
The DNNs falsely recognize the input data contaminated with such perturbation and it leads to mislabeling, mis-segmentation of medical images \cite{finlayson2019adversarial} as well as road signs \cite{eykholt2018robust}, and pedestrians \cite{hendrik2017universal}.
Consequently, this problem greatly affects the reliability of most systems developed using DNNs, which are suitable for critical applications such as medical imaging, autonomous car, and so on.

\begin{figure}[h]
    \centering  
    \includegraphics[width=85mm]{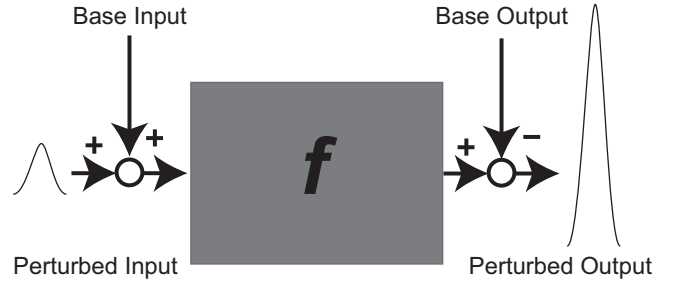}
    \caption{The concept of the adversarial perturbation generated in the present study; $\bm{f}$ represents an input-output relationship of a DNN; the perturbation superimposed on a base input is directly outputted on being superimposed on a base output.}
    \label{fig:concept}  
\end{figure}
Randomly generated perturbations superimposed on an input for a DNN are likely to be attenuated inside the DNN and do not have a significant effect on an output when the magnitude of the perturbation is low.
As shown in Fig.~\ref{fig:concept}, inputted perturbation, which penetrates a DNN, and appears in the output of the DNN is, considered to be a particular threat even when the magnitude of the perturbation is the same.
Thus, it is challenging to prevent such noise from outputting erroneous results.
In this study, we show the existence of such perturbations mathematically.
The present study aims at seeking such input perturbation whose magnitude is not attenuated but amplified and then penetrates a DNN without knowing the inner states of the DNN.

Compared with previous studies, our method has a singularly distinct combination of characteristics valuable for attackers as follows:
\begin{itemize}
    \item \textbf{Black-Box Attack} -- The proposed method does not use the inner state of the DNNs as in white-box attacks, such as the utilization of a backpropagation \cite{papernot2016limitations}, and monitoring inner layer variables \cite{khrulkov2018art}, but only request an input-output relationship, i.e., an oracle.
    \item \textbf{Query Efficient} -- The previous methods require lots of black-box evaluation when limited to black-box attacks, and its computational cost is typically $\mathcal{O}\left( h\cdot w\right)$ where $h$ and $w$ represent the height and width of input images, respectively. However, the proposed method achieves $\mathcal{O}\left(1\right)$ complexity.
    \item \textbf{Memory Efficient} -- Unlike other gradient methods applied in white-box attacks in which the construction of the Jacobian matrix is explicitly required, the proposed method requires an implicit expression of the matrix only, which reduces the memory requirements to $\mathcal{O}\left( h \cdot w\right)$ from $\mathcal{O}\left( h^2 \cdot w^2\right)$.
    \item \textbf{Generality} -- The proposed method does not assume components of an input vector for DNNs; hence, the method can be applied to DNNs for voice, text, and video recognition, other than image processing only.
\end{itemize}

\section{Related works}
There are three streams of research for the generation of adversarial perturbation.
\begin{figure}[h]
    \centering
    \includegraphics[width=90mm]{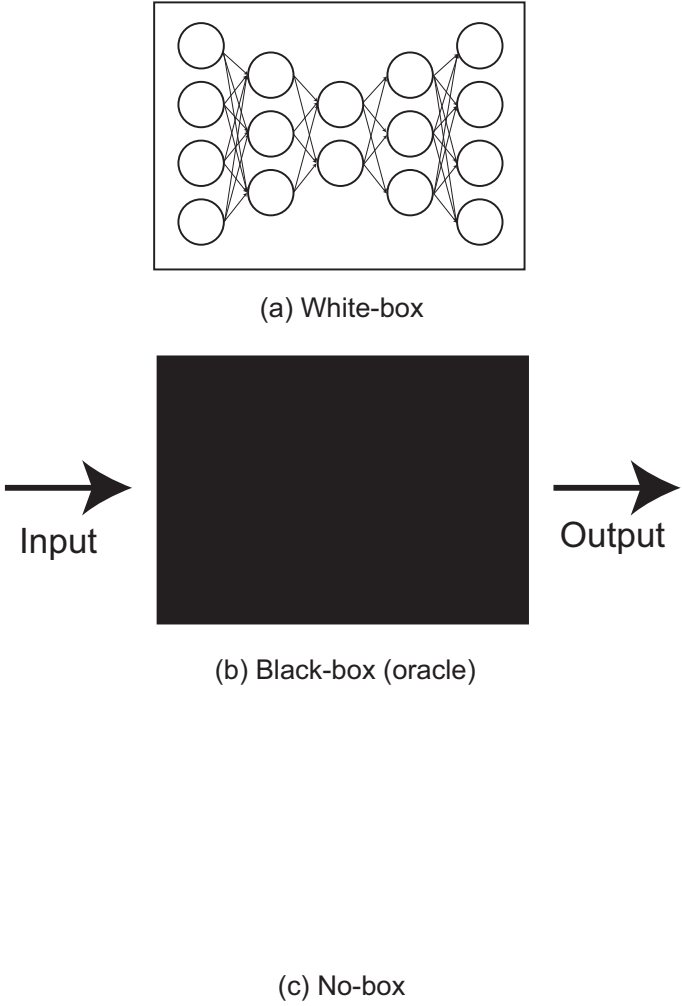}
    \caption{Definitions of white-box, black-box with an oracle, and no-box model; in the black-box with an oracle, input-output relationship is only available and the inner states of a DNN are completely concealed.}
    \label{fig:kind}    
\end{figure}
The first one is white-box attacks \cite{papernot2016limitations}.
In white-box attacks, the inner states of the DNNs must be visible and/or the inner application programming interfaces (APIs) of the DNNs (e.g., a function to compute gradients of any output vector using backpropagation technique), must be accessible from attackers (see Fig.~\ref{fig:kind}(a)).
Next, the second one is the black-box attacks \cite{su2019one,chen2017zoo,tramer2017space,moosavi2017universal}.
In black-box attacks, there is no access permission for those in white-box attacks but an oracle only (see Fig.~\ref{fig:kind}(b)).
Further, the last one is the no-box attacks \cite{ hendrik2017universal, khrulkov2018art,moosavi2017universal}, which supports the DNNs attack where the oracle is not reachable (see Fig.~\ref{fig:kind}(c)).
In this case, attackers design perturbations, which have a universal impact on several DNNs.
Note that the research works investigated in this section are limited to black-box attacks since the present study focuses only on black-box attacks, which is considered to be the most realistic situation for attacks against the DNNs.

\begin{table}[htb]
\caption{Classification of related works with oracle evaluation.}
\centering
  \begin{tabular}{|c|c|c|c|} \hline
    Method & Number of oracle evaluation & Memory requirement \\ \hline  
    Present(ours) & $\mathcal{O}(1)$&$\mathcal{O}(h\cdot w)$\\ \hline
    One pixel \cite{su2019one}& $\mathcal{O}(h\cdot w)$& $\mathcal{O}(h\cdot w)$  \\ \hline
    ZOO \cite{chen2017zoo} & $\mathcal{O}(h\cdot w)$ & $\mathcal{O}(h^2\cdot w^2)$ \\ \hline
  \end{tabular}
  \label{tbl:performance}
\end{table}
Table~\ref{tbl:performance} summarizes the computational complexity for oracle evaluation and memory requirement needed to execute algorithms for three different black-box attacks based on oracle evaluation.

One-pixel attack \cite{su2019one} only adds perturbation onto one pixel in an input image, and the perturbation is optimized using the concept of differential evolution.
Note that in a classification problem introduced in \cite{su2019one}, the dimension of an output of an oracle evaluated is much less than that of input.

In attacks using zeroth-order optimization (ZOO) \cite{chen2017zoo}, the gradient of an output of an oracle ($\bm{f}$) with reference to the $i$-th component of an input vector ($\bar{x}_i$), i.e., $i$-th column vector of the Jacobian matrix, is explicitly computed using the central differential formula defined as:
\begin{equation}
    \label{eqn:numerical_diff}
    \frac{\partial \bm{f}}{\partial x_i} \approx \frac{\bm{f}\left( \bar{\bm{x}} + \epsilon \cdot \bm{e}_i\right)- \bm{f}\left( \bar{\bm{x}} - \epsilon\cdot \bm{e}_i\right)}{2\epsilon},
\end{equation}
where $\bm{e}_i$ is a unit vector whose $i$-th component is 1 and $\epsilon$ is a sufficiently small real constant used in the numerical differentiation.
Therefore, the computational cost of this method becomes unrealistic if the dimension of input vector becomes larger.
Note that the above differential formula computes the component of the Jacobian matrix directly, whereas our method calculates the Jacobian matrix-vector multiplication instead.

\section{Algorithm}
Consider an oracle of a DNN, which maps a variable vector of $n$ dimension such as an image of street scape ($\bm{x} \in \mathbb{R}^n$) onto another variable vector of $n$ dimension such as a segmented image.
In this study, $n$ equals to $h\cdot w$ since images are dealt with.
For simplicity, this mapping is denoted as $\bm{f}(\cdot)$.

Fig.~\ref{fig:concept} shows a characteristic of the perturbation generated in our study.
Our primary interest is to seek a perturbation, which is not attenuated by the function $\bm{f}\left(\cdot \right)$, and potentially penetrates a DNN within the extent of linearity.

Consider to add a unit perturbation vector $\tilde{\bm{x}}$ multiplied with the small coefficient $\epsilon$, to a base image vector $\bar{\bm{x}}$.
Then, the function can be expanded with Taylor series as follows:
\begin{equation}
\label{eqn:linearex}
    \bm{f}\left(\bar{\bm{x}} + \epsilon\tilde{\bm{x}}\right) = \bm{f}\left(\bar{\bm{x}}\right) + \epsilon \mathcal{J} \tilde{\bm{x}} + \mathcal{O}\left(\epsilon^2\right),
\end{equation}
and
\begin{equation}
\label{eqn:linearex_minus}
    \bm{f}\left(\bar{\bm{x}} - \epsilon\tilde{\bm{x}}\right) = \bm{f}\left(\bar{\bm{x}}\right) - \epsilon \mathcal{J} \tilde{\bm{x}} + \mathcal{O}\left(\epsilon^2\right),
\end{equation}
where $\mathcal{J} := \left(\frac{\partial\bm{f}\left(\bm{x}\right)}{\partial \bm{x}}\right)_{\bm{x} = \bar{\bm{x}}} \in \mathbb{R}^{n\times n}$ is the Jacobian matrix of the oracle.

To generate perturbations ($\bm{v}_i$) such that all the components inputted in an oracle can be outputted with being multiplied by a constant, the following relationship must be satisfied:
\begin{equation}
    \mathcal{J} \bm{v}_i = \lambda_i \bm{v}_i,
\end{equation}
where $\lambda_i$ is the constant (an eigenvalue) and $\bm{v}_i$ is the perturbation (an eigenvector).
We assume these eigenvalues are sorted in descending order based on magnitudes, i.e., the diagonal components are sorted such that $|\lambda_{1}| \ge |\lambda_{2}| \ge \ldots \ge |\lambda_{n}|$.
These perturbations are named \textit{penetrative perturbation}.

By inputting the eigenvector, an output of an oracle is obtained by modifying \eqref{eqn:linearex} as:
\begin{equation}
    \label{eqn:explicit}
    \bm{f}\left(\bar{\bm{x}} + \delta\cdot{\bm{v}}_i \right) = \bm{f}\left(\bar{\bm{x}}\right) + \delta\cdot \lambda_i {\bm{v}}_i +  \mathcal{O}\left(\delta^2\right),
\end{equation}
where $\epsilon$ is replaced with $\delta$ to distinguish the one used in numerical differentiation and the other used in adversarial attacks.

To extract the eigenvalue with the largest magnitude together with the corresponding eigenvector from $\mathcal{J}$, ARPACK \cite{lehoucq1998arpack}, which is an implementation of the implicitly restarted Arnoldi method (IRAM), is applied.

In IRAM, same as other eigenvalue problem solvers, a mapping from $\tilde{\bm{x}}$ onto $\mathcal{J}\tilde{\bm{x}}$ must be defined.
However, as pointed out in the previous study using the Jacobian matrix explicitly as an implementation technique for Jacobian-based Saliency Map Attack (JSMA) \cite{papernot2016limitations}, the construction of the Jacobian matrix is time-consuming.
To explicitly generate the Jacobian matrix, multiple evaluations ($n+1$ times with forward difference, and $2n$ times with central difference) of the function $\bm{f}\left(\cdot \right)$ is necessary.
In this study, we applied the Fr\'{e}chet derivative to reduce the evaluation time by implicitly forming the Jacobian matrix with the following approximation:
\begin{equation}
    \label{eqn:Jx}
    \mathcal{J} \tilde{\bm{x}} = \frac{\bm{f}\left(\bar{\bm{x}} + \epsilon \cdot \tilde{\bm{x}}\right) - \bm{f}\left(\bar{\bm{x}} - \epsilon \cdot \tilde{\bm{x}} \right) }{2\epsilon} + \mathcal{O}\left(\epsilon^3\right).
\end{equation}
Note that there is a crucial difference between \ref{eqn:Jx} and \eqref{eqn:numerical_diff}.
With \ref{eqn:Jx}, the mapping from $\tilde{\bm{x}}$ onto $\mathcal{J}\tilde{\bm{x}}$ can be computed without knowing the components of the Jacobian matrix.
The algorithm to compute the mapping is given as follows: 
\begin{algorithm}[H]
\caption{The Jacobian matrix-vector multiplication}
\begin{algorithmic}[1]
    \REQUIRE{$\bm{f}, \tilde{\bm{x}}, \bar{\bm{x}}, \epsilon$}
    \ENSURE{$\mathcal{J}\tilde{\bm{x}}$}
    \STATE{Evaluate a DNN: $\bm{f}_+ \leftarrow \bm{f} \left( \bar{\bm{x}} + \epsilon \cdot \tilde{\bm{x}} \right)$}
    \STATE{Evaluate a DNN: $\bm{f}_- \leftarrow \bm{f} \left( \bar{\bm{x}} - \epsilon \cdot \tilde{\bm{x}} \right)$}        
    \STATE{Fr\'{e}chet derivative: $\mathcal{J}\tilde{\bm{x}} \leftarrow \frac{\bm{f}_+ - \bm{f}_-}{2\epsilon}$}
    \RETURN{$\mathcal{J}\tilde{\bm{x}}$}
\end{algorithmic}
\end{algorithm}

A typical DNN accepts only a real vector as an input; hence, the input vector ($\tilde{\bm{x}}$) must have only real values, which indicates that an eigenvalue problem solver for symmetric matrix (e.g., \texttt{eigsh} in SciPy package) is only executable.
Consequently, the algorithm to generate the penetrative perturbation with ARPACK is written as follows:
\begin{algorithm}[H]
\caption{Perturbation Generator using ARPACK}
\begin{algorithmic}[1]
    \REQUIRE{$\bm{f}, \bar{\bm{x}}, \epsilon, \mathrm{tol}, \mathrm{itmax}$}
    \ENSURE{$\bm{v}_1, \lambda_1$}
    \STATE{Prepare linear operator (Algorithm 1)}
    \STATE{Invoke \texttt{eigsh} with \texttt{LM} option, given the convergence tolerance ($\mathrm{tol}$), maximum number of Arnoldi iteration ($\mathrm{itmax}$), and the operator.}
\end{algorithmic}
\end{algorithm}
One of the \texttt{eigsh} methods defined in SciPy package accepts the functor in which the Jacobian matrix-vector multiplication is defined to compute eigenvalues and corresponding eigenvectors of the Jacobian matrix.
With the \texttt{LM} option, the solver finds the eigenvalues from those with the largest magnitude. 
Note that the Jacobian matrix of a DNN is generally non-symmetric; hence, the convergence of eigenvalues may be affected.

\section{Numerical experiments and results}
\label{sec:result}
\begin{figure}
    \centering
    \includegraphics[width=85mm]{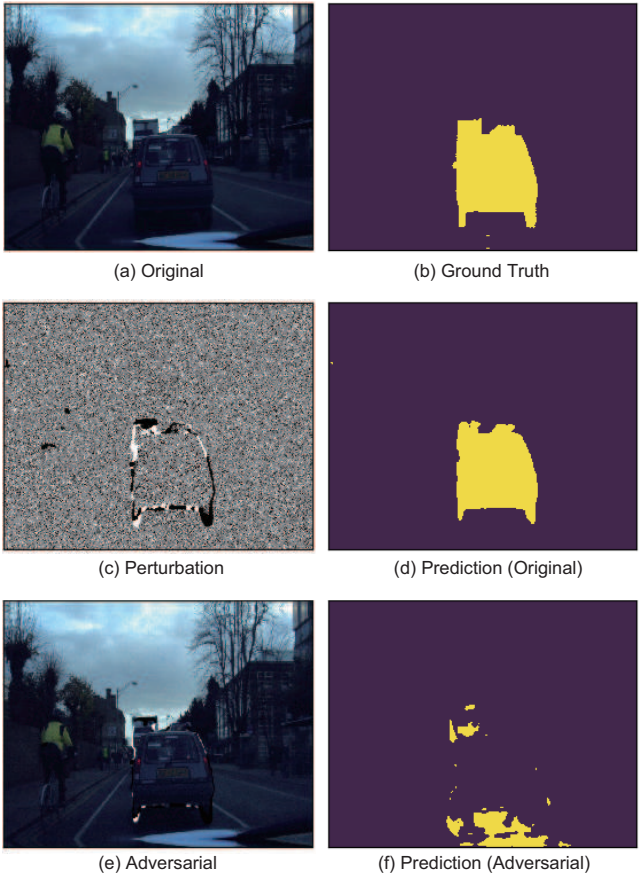}
    \caption{Segmentation result; (a) original image inputted to the DNN, (b) ground truth of the segmented image, (c) perturbation to be added to the original image, (d) prediction result for the original image, (e) linear combination of the original image and the perturbation (adversarial image), and (f) prediction result for the adversarial image.}
    \label{fig:segmentation}
\end{figure}

In this section, by only knowing an oracle concerning a DNN, we performed a numerical experiment to confirm the applicability of the proposed algorithm and the effectiveness of the penetrative perturbation. 
As a target network, a DNN for segmentation is chosen.
Thus, a DNN, in which UNet is employed as a model, and EfficientNet-b3 is adopted as a backbone \cite{DBLP:journals/corr/abs-1905-11946, Yakubovskiy:2019}, is selected as one of the state-of-the-art networks to execute semantic segmentation tasks.
The network is trained to segment cars in given images with using Adam optimizer to minimize the summation of Focal and Dice loss \cite{Yakubovskiy:2019}.
The dataset for training and testing the DNN is taken from CamVid \cite{BrostowSFC:ECCV08}, which contains 367 and 233 images for both training and testing, respectively.
The number of output channel is 1, whereas that of the input channel is 3; hence, there is a degree of freedom to handle a perturbation input superimposed on an original image.
Therefore, the generated perturbation is equally inputted to all the input channels, which potentially affects the convergence of the algorithm.
In this experiment, $\epsilon$, $\delta$ are set to $10^{-4}$ and $2\times 10^{-3} \cdot n$, respectively.
During the iteration procedure using the \texttt{eigsh}, the output of the oracle, i.e., a segmentation result, is not rounded to the nearest value.
Note that the prediction results shown in Fig.~\ref{fig:segmentation} are rounded to 0 or 1.
In this numerical experiment, $n=184,320$; hence, the number of the oracle evaluation is 368,640 times when one tries to construct the Jacobian matrix explicitly using \eqref{eqn:numerical_diff}.
The versions of Python and SciPy are set to 3.6.5 and 1.1.0, respectively.

The eigenvalue with the largest magnitude converged typically within 2 Arnoldi iterations and with 39 oracle evaluations on average when $\mathrm{tol} = 10^{-12}$.
The real part of eigenvalues with the largest magnitude out of 233 images is distributed from -30.5 to 28.3.
This means that the magnitude of an input perturbation can be magnified 30 times approximately, and appears in an output.
The eigenvector corresponding to the eigenvalue with the largest magnitude extracted is superimposed on the input vector, and the segmentation is executed.   
Fig.~\ref{fig:segmentation} shows the result.
Although the perturbation added is little to avoid contaminating the original image globally, the segmented image significantly collapsed, and most portion ought to be recognized as cars are recognized as void.

To quantitatively evaluate the performances of the proposed method, we conducted statistical analysis.
Moreover, we generated perturbed images with uniformly distributed random noise of the same magnitude, and performed semantic segmentation.

To evaluate the similarity between an original image and an adversarial image, we selected Structural SIMilarity (SSIM) index \cite{wang2004image}.
Also, to evaluate the performance of the segmentation, we adopted the Dice coefficient.
\begin{table}[htb]
\caption{Effects of penetrative perturbation against EfficientNet-B3. Mean values of SSIM and Dice coefficient are shown with variances indicated within brackets.}
\centering
  \begin{tabular}{|l|ll|} \hline
            & SSIM & Dice  \\ \hline  
    Uniform & 0.106 (0.00220) & 0.829 (0.0489) \\
    Penetrative ($\mathrm{tol} = 10^{-3}$) & 0.861 (0.0325) & 0.254 (0.110) \\ 
    Penetrative ($\mathrm{tol} = 10^{-6}$) & 0.861 (0.0325) & 0.259 (0.109) \\ 
    Penetrative ($\mathrm{tol} = 10^{-12}$) & 0.860 (0.0332) & 0.251 (0.110) \\ \hline
  \end{tabular}
  \label{tbl:statistical}
\end{table}

Table~\ref{tbl:statistical} shows the statistical results obtained.
The mean values of SSIM with the penetrative perturbation are higher compared to uniformly distributed random noise.
The difference shows, irrespective of the same magnitude, the penetrative perturbation is less likely to destroy the structure of images than using uniformly distributed random perturbations.
The mean values of the Dice coefficient with the penetrative perturbation are less than half compared with uniform perturbation.
Moreover, variances of Dice coefficient with the penetrative perturbation are approximately twice compared the other perturbation. 
Consequently, the uniformly distributed random perturbations constantly fail to deceive the DNN, whereas the penetrative perturbation can deceive the DNN with higher uncertainty.
Additionally, setting the convergence tolerance to $10^{-3}$ could affect the results slightly.

\section{Discussion}
\subsection{Distribution characteristics of the perturbation}
The perturbation generated is mainly distributed around target objects to be segmented.
Hence, the proposed algorithm has a mechanism to make object boundary obscure or accented for the DNN to induce mis-segmentation.
\begin{table}[htb]
    \caption{Effects of penetrative perturbation against EfficientNets. The mean values of SSIM and Dice coefficient are shown with variances indicated within brackets.}
    \centering
    \begin{tabular}{|l|ll|} \hline
            & SSIM & Dice  \\ \hline
    Uniform (B0) & 0.106 (0.00220) & 0.747 (0.0766) \\              
    Penetrative (B0, $\mathrm{tol} = 10^{-12}$) &0.870 (0.0205) & 0.436 (0.108) \\  
    Uniform (B1) & 0.106 (0.00220) & 0.829 (0.0410) \\ 
    Penetrative (B1, $\mathrm{tol} = 10^{-12}$) & 0.869 (0.0276) & 0.343 (0.0764) \\    
    Uniform (B2) & 0.106 (0.00220) & 0.797 (0.0571) \\ 
    Penetrative (B2, $\mathrm{tol} = 10^{-12}$) & 0.872 (0.0223) & 0.250 (0.0473) \\
    Uniform (B3) & 0.106 (0.00220) & 0.829 (0.0489) \\    
    Penetrative (B3, $\mathrm{tol} = 10^{-12}$) & 0.861 (0.0325) & 0.254 (0.110) \\
    Uniform (B4) & 0.106 (0.00220) & 0.823 (0.0497) \\ 
    Penetrative (B4, $\mathrm{tol} = 10^{-12}$) & 0.870 (0.0183) & 0.326 (0.0981) \\    \hline
    \end{tabular}
    \label{tbl:depth}  
\end{table}

There is a possibility that the characteristics of the perturbation are affected depending on tasks built upon different DNN architecture and dataset.
Hence, we investigated the effects of the penetrative perturbation against different EfficientNets, as shown in Table~\ref{tbl:depth}.
The DNNs are constantly deceived according to the Dice coefficients with the penetrative perturbation compared with uniformly distributed noise.   

\subsection{Limitation of the perturbation}
\begin{figure}
    \centering
    \includegraphics[width=85mm]{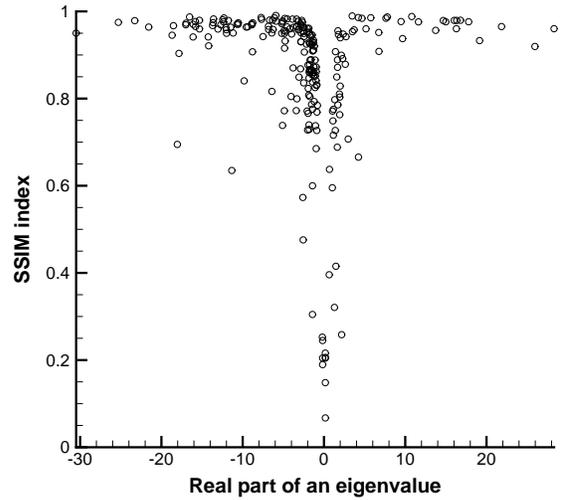}
    \caption{Correlation between the real part of eigenvalue and SSIM index.}
    \label{fig:correlation_ssim}
\end{figure}
Fig.~\ref{fig:correlation_ssim} shows the correlation between the real part of eigenvalues with the largest magnitude and the SSIM indices for each of the test images.
There is a tendency that the SSIM index corresponding to eigenvalues with a magnitude less than 2, approximately, is lower among the others.
Therefore, there is a certain image set with which the DNN is not deceived using the penetrative perturbation without significantly modifying an original image.

\subsection{Non-linearity effects for the perturbation}
From (\ref{eqn:explicit}), the superimposed penetrative perturbation appears on the base output vector as it is within the extent of linearity.
In this study, the unit eigenvector computed, is multiplied by a constant $\delta$, so that the perturbation can have significant effects in the segmentation task.
As discussed in Fig.~\ref{fig:segmentation}, the perturbation does not appear on the output as it is.
The first reason for this is that the linearity of the perturbation is not guaranteed with the parameter ($\delta = 2.0 \times 10^{-3} \cdot n$).
The other possible reason is that the sigmoid activation functions deployed in the output layer could limit the output.

\subsection{Behavior of higher eigenvectors}
To simplify the problem, we assume real values for all the eigenvalues.
Based on our earlier formulation, eigenvectors corresponding with the eigenvalues with the second largest magnitude or higher have less impact as an adversarial perturbation due to (\ref{eqn:explicit}).
However, the behavior of such higher eigenvectors over semantic segmentation is not apparent.
Therefore, the higher eigenvectors are extracted and the same analysis as in Section~\ref{sec:result} is conducted.
\begin{figure}
    \centering
    \includegraphics[width=85mm]{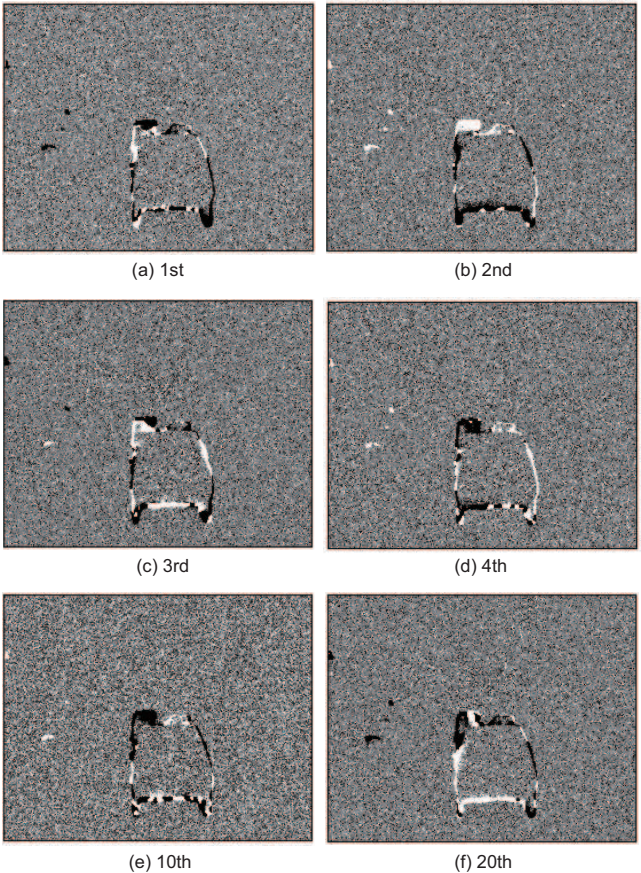}
    \caption{Eigenvectors extracted; (a) the eigenvector corresponding with the eigenvalue with the largest magnitude ($\lambda = -14.5$), (b) with the second largest magnitude ($\lambda = -11.7$), (c) with the third largest magnitude ($\lambda = -10.2$), (d) with the fourth largest magnitude ($\lambda = 9.53$), (e) with the 10th largest magnitude ($\lambda = -6.87$), (f) with the 20th largest magnitude ($\lambda = -5.62$).}
    \label{fig:modes}
\end{figure}

\begin{table}[htb]
\caption{Effects of higher penetrative perturbation against EfficientNet-B3. Mean values of SSIM and Dice coefficient are shown with variances indicated within brackets. All the computation is conducted with $\mathrm{tol} = 10^{-12}$.}
\centering
    \begin{tabular}{|l|ll|} \hline
            & SSIM & Dice  \\ \hline
    Penetrative (1st) & 0.860 (0.0332) & 0.251 (0.110) \\
    Penetrative (2nd) & 0.838 (0.0366) & 0.201 (0.0875) \\ 
    Penetrative (3rd) & 0.822 (0.0369) & 0.145 (0.0581) \\
    Penetrative (4th) & 0.799 (0.0406) & 0.129 (0.0491) \\   
    Penetrative (10th) & 0.707 (0.0578) & 0.109 (0.0435) \\    
    Penetrative (20th) & 0.590 (0.0704) & 0.0949 (0.0387)\\ \hline    
    \end{tabular}
    \label{tbl:modes}  
\end{table}

Fig.~\ref{fig:modes} shows the visualization result of the higher eigenvectors, and the result of the statistical analysis is shown in Table~\ref{tbl:modes}.
As stated earlier in our numerical experiment, $n=184,320$; the 20th eigenvector is still considered predominant, among others.
The perturbation expressed with higher eigenvectors is distributed around an object to be segmented, and this is the same tendency as that of the first eigenvector.
Moreover, the perturbation based on a higher eigenvector can destroy original images and segmentation results more significantly than those of the first eigenmode vector according to the SSIM and Dice coefficient as shown in Table~\ref{tbl:modes}.

Finally, it is possible to blend these eigenvectors linearly in order not to give unusual impact on human eyes with adversarial characteristics being kept as follows:
\begin{equation}
    \tilde{\bm{x}} = \sum_{i=1}^{M} a_i \bm{v}_i,
\end{equation}
where $a_i \in \mathbb{R}$ is a constant arbitrary chosen, and $M$ is the number of the predominant eigenvectors taken.

\subsection{Existence of complex eigenvalues and eigenvectors}
All the eigenvalues are real numbers if the Jacobian matrix is symmetric.
In realistic DNNs, in general, the matrix takes non-symmetric form.
Therefore, a complex eigenvalue is potentially selected as $\lambda_1$ if one strictly solves the system.
However, with a DNN where an oracle evaluation is only permitted, it is realistic to assume that it is not allowed to add complex perturbation on an original vector such as an image.
Practically, it is possible to approximately solve the eigenvalue problem for the non-symmetric Jacobian matrix, as presented in this study.

\section{Conclusions and Future works}
We proposed a method to compute the penetrative perturbation based on ARPACK as an eigenvalue problem solver and Fr\'{e}chet derivative as a generator of the Jacobian matrix-vector product.
The method proves a smaller number of query trials and fewer memory requirements among other black-box attack algorithms.
From numerical evaluation, with this perturbation, one of the state-of-the-art DNNs, with which high-performance segmentation is possible, can be deceived with the realistic number of query trials.

In future works, we expect to achieve the following: (i) to establish a method to systematically detect the penetrative perturbations introduced in our study as a defender and (ii) to reveal how the penetrative perturbation propagates inside a DNN with considering nonlinear factors.

\section*{Acknowledgment}
The Department of Computational Diagnostic Radiology and Preventive Medicine, The University of Tokyo Hospital, is sponsored by HIMEDIC Inc. and Siemens Healthcare K.K.

\ifCLASSOPTIONcaptionsoff
  \newpage
\fi



\bibliographystyle{IEEEtran}
\bibliography{main.bib}
\end{document}